%% file: main_IEEE-conference-template-062824.tex
\documentclass[conference]{IEEEtran}
\IEEEoverridecommandlockouts
\usepackage{cite}
\usepackage{amsmath,amssymb,amsfonts}
\usepackage{algorithm}
\usepackage{algorithmic}
\usepackage{graphicx}
\usepackage{textcomp}
\usepackage{xcolor}
\usepackage{setspace}
\usepackage{subcaption}
\usepackage{multirow}
\usepackage{comment}
\usepackage{stfloats}
\usepackage{orcidlink}
\usepackage{wrapfig}
\usepackage{multicol} % Add this to your preamble
\usepackage{afterpage}
\usepackage{wrapfig}  % This package is for wrapping text around figures

\def\BibTeX{{\rm B\kern-.05em{\sc i\kern-.025em b}\kern-.08em
    T\kern-.1667em\lower.7ex\hbox{E}\kern-.125emX}}
\begin{document}

\title{Personalizing Federated Learning for Hierarchical
Edge Networks with Non-IID Data}

\author{
    \IEEEauthorblockN{
    Seunghyun Lee\IEEEauthorrefmark{1}\orcidlink{0000-0003-4343-6931}, 
    Omid Tavallaie\IEEEauthorrefmark{2}\orcidlink{0000-0002-3367-1236}, 
    Shuaijun Chen\IEEEauthorrefmark{1}\orcidlink{0009-0001-4944-3406}, 
    Kanchana Thilakarathna\IEEEauthorrefmark{1}\orcidlink{0000-0003-4332-0082}, \\
    Suranga Seneviratne\IEEEauthorrefmark{1}\orcidlink{0000-0002-5485-5595},
    Adel Nadjaran Toosi \IEEEauthorrefmark{3}\orcidlink{0000-0001-5655-5337},
    Albert Y. Zomaya \IEEEauthorrefmark{1}\orcidlink{0000-0002-3090-1059}}
    \IEEEauthorblockA{\IEEEauthorrefmark{1} School of Computer Science, The University of Sydney, Australia}
    \IEEEauthorblockA{\IEEEauthorrefmark{2} Department of Engineering Science, University of Oxford, United Kingdom}
    \IEEEauthorblockA{\IEEEauthorrefmark{3} School of Computing and Information Systems, The University of Melbourne, Australia}
}
\maketitle

\begin{abstract}
Accommodating edge networks between IoT devices and the cloud server in Hierarchical Federated Learning (HFL) enhances communication efficiency without compromising data privacy. However, devices connected to the same edge often share geographic or contextual similarities, leading to varying edge-level data heterogeneity with different subsets of labels per edge, on top of  device-level heterogeneity. This hierarchical non-Independent and Identically Distributed (non-IID) nature, which implies that each edge has its own optimization goal, has been overlooked in HFL research. Therefore, existing edge-accommodated HFL demonstrates inconsistent performance across edges in various hierarchical non-IID scenarios. To ensure robust performance with diverse edge-level non-IID data, we propose a Personalized Hierarchical Edge-enabled Federated Learning (PHE-FL), which personalizes each edge model to perform well on the unique class distributions specific to each edge. We evaluated PHE-FL across 4 scenarios with varying levels of edge-level non-IIDness, with extreme IoT device level non-IIDness. To accurately assess the effectiveness of our personalization approach, we deployed test sets on each edge server instead of the cloud server, and used both balanced and imbalanced test sets. Extensive experiments show that PHE-FL achieves up to 83\% higher accuracy compared to existing federated learning approaches that incorporate edge networks, given the same number of training rounds. Moreover, PHE-FL exhibits improved stability, as evidenced by reduced accuracy fluctuations relative to the state-of-the-art FedAvg with two-level (edge and cloud) aggregation.
\end{abstract}


\input{introduction.tex}
\input{background.tex}
\input{PHEFL.tex}
\input{experiment.tex}
\input{result.tex}

\input{related_works.tex}
\input{conclusion.tex}

\bibliographystyle{IEEEtran}
\bibliography{references}

\end{document}

%% file: introduction.tex
\section{Introduction}

Federated Learning (FL) is an emerging Machine Learning (ML) framework that achieves high accuracy without requiring the sharing of local data with a centralized server. Involving IoT devices and a central cloud server, 2-level FL aggregation framework was first proposed under the name FederatedAveraging (FedAvg) algorithm \cite{mcmahan2017}. In FedAvg, IoT devices train models individually and then transmit the model weights to the cloud server. The server then averages these weights to create an aggregated global model that performs well and therefore can be deployed across all participating devices. Given that IoT devices often handle privacy-sensitive data, the ability to retain data locally while still contributing to a high-performing global model has gained significant interest from both academia and industry. Leading companies like Meta \cite{greenfl} and Google \cite{mcmahan2017} have been at the forefront of FL research, implementing various FL techniques in large-scale real-world testbeds and integrating them into their products.

Although IoT devices benefit from not having to transmit local data to the cloud while having an aggregated model from the server that performs well on itself, several challenges have emerged. These challenges include vulnerability to inversion attacks \cite{InvertingGradients,ModelInversionAttack} and high communication costs \cite{greenfl,CommunicationEfficiency}, and earlier implementations of 2-level FL often required trade-offs between security and communication efficiency \cite{SecureAggregation,tradeoff-2}. However, recent advancements have introduced a 3-level FL architecture \cite{3-levelFL,wentao} that effectively addresses both issues. In 3-level FL, which we named as \textbf{EdgeCloud}, device models are first aggregated
at the edge, then at the cloud, and the global model is
uniformly distributed to all clients. 
Integrating an intermediary edge layer into the conventional 2-level system not only enhances data privacy by restricting direct access to IoT device models from the cloud server but also reduces communication costs by enabling model transmission from IoT devices to the nearby edge server instead of the cloud server, without using the internet \cite{Edge-FL-survey,survey-2}.

\begin{figure}
    \centering
        \includegraphics[width=8cm]{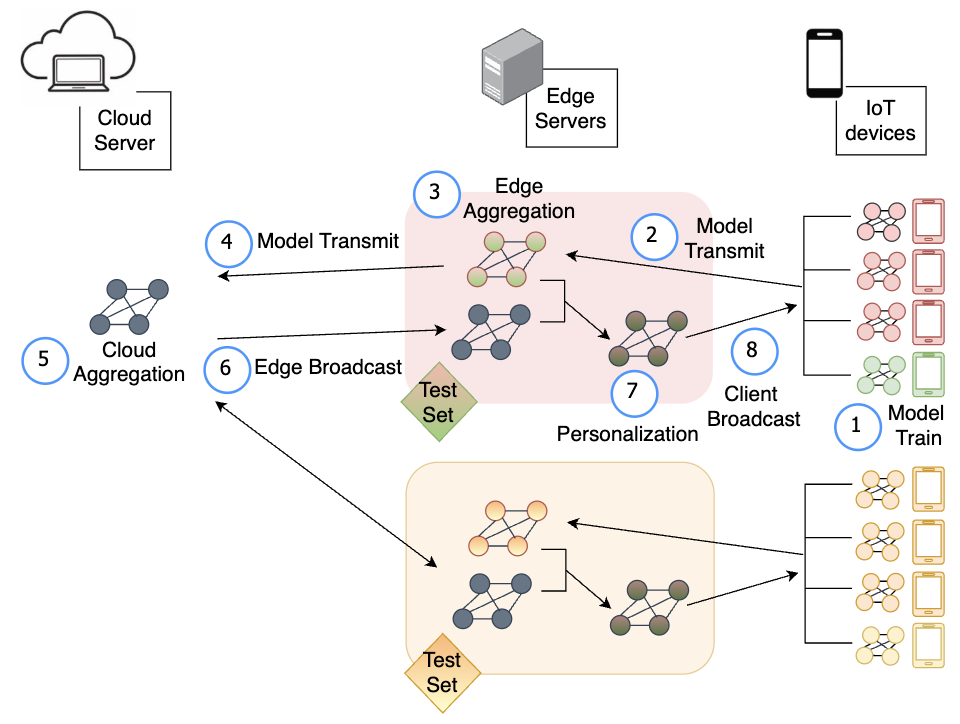}
    \caption{Personalized Hierarchical Edge-enabled Federated Learning (\textbf{PHE-FL}). After cloud aggregation, PHE-FL creates personalized models at the edge, allowing connected clients to receive distinct, edge-personalized models.}
    \label{fig:PHE-FL}
    \vspace{-10pt}
\end{figure}

Data heterogeneity, also known as non-IIDness (Independent and Identically Distributed), is another significant issue in FL. Traditional 2-level FL methods often perform poorly with non-IID data \cite{FairFL} because the global model may diverge from the optimal local models fitted to specific client data distributions. Many studies \cite{shuaijun,Feed,measuring-non-iid,HPFL,HCPFL} have explored strategies within the 2-level FL framework to address non-IIDness.

In 3-level FL, which complements 2-level FL by enhancing privacy and reducing communication costs, data heterogeneity remains a major challenge, limiting its real-world application. This is due to \textbf{hierarchical non-IIDness}, where non-IID data exists at both the device and edge levels. Devices often show extreme non-IID data distributions with minimal overlap \cite{deviceNonIID,deviceNonIID2}, while edges experience varying degrees of heterogeneity influenced by geographic and contextual proximity \cite{EdgeNetwork}. For example, in smart city traffic or wildlife monitoring, devices connected to the same edge may collect relatively homogeneous data, but edge-wise, each edge in different region monitor distinct subsets of labels (e.g., traffic types or species). This results in \textbf{different optimization goals for each edge}, and relying on a single global model causes performance degradation due to misalignment with edge-specific objectives, yet this issue remains inadequately addressed in existing research.

To address hierarchical non-IIDness and improve 3-level FL for real-world applications, where each edge has a distinct optimization goal, we developed \textbf{Personalized Hierarchical Edge-based Federated Learning (PHE-FL)}. PHE-FL personalizes models at each edge by identifying the most relevant model between the \textit{edge-aggregated model} of the target edge and the \textit{cloud-aggregated model}, which is the aggregation of all edges except the target. The more relevant model is assigned a greater weight in the target edge’s personalized model. To evaluate the effectiveness of PHE-FL for edges with distinct optimization goals, we deploy test sets on edge servers instead of the cloud server. Our results show that PHE-FL maintains high accuracy and consistent performance across various edge-level non-IID scenarios, demonstrating its robustness compared to other 3-level FL architectures.

In summary, our research presents 3 key contributions:
\begin{itemize}
  \item[\textbullet] We have introduced a new concept known as hierarchical non-IIDness of data. To the best of our knowledge, no existing research in the field of 3-level FL has explicitly explored this concept. We have identified 4 scenarios characterized by different degrees of edge-level non-IIDness with label skewness and conducted extensive experiments within these settings.

  \item[\textbullet] We have developed a novel method, PHE-FL, for personalizing 3-level FL models at the edge level without compromising data privacy or causing additional communication cost. This method generates a unique personalized model for each edge in every round. PHE-FL is independent of data distribution and application, making it suitable for any image and potentially any classification problem. 
  
  \item[\textbullet] Through extensive experiments, we demonstrate that PHE-FL achieves high accuracy and maintains strong robustness among edge-accommodated FL approaches under various degrees of edge-level non-IIDness.
\end{itemize}

%% file: background.tex
\section{Background}
Specific notations used throughout the paper are listed in Table  \ref{tab:notation}. The terms \textit{client} and \textit{IoT device} have been used interchangeably throughout the paper.
\begin{table}
\centering
\scriptsize 
\caption{Notation Summary}
\begin{tabular}{ p{2cm}  p{6cm} }
\hline
\textbf{Notation} & \textbf{Description} \\ \hline
$GM$ & Global Cloud Server Model \\ \hline
$CAM^t_i $& Cloud Aggregated Model for edge ${i}$ in the ${t}$-th round \\ \hline
$EAM_i^t$ & Edge Aggregated Model for edge ${i}$ in the ${t}$-th round \\ \hline
$PEAM_i^t$& Personalized Edge Aggregated Model for edge ${i}$ in the ${t}$-th round  \\ \hline
$DM^t_i$ & Device Model for IoT device ${i}$ in the ${t}$-th round  \\ \hline
$PDM_i^t$& Personalized Device Model for IoT device ${i}$ in the ${t}$-th round  \\ \hline
$K$ & set of edges \\ \hline 
$E_i$ & subset of clients that belongs to Edge ${i}$ \\ \hline
$C$ & set of clients \\ \hline
$TTD_i$ & Total Test Set for edge $i$ \\ \hline
$PTD_i$ & 15\% of $TTD_i$ for personalization \\ \hline
$ETD_i$ & 85\% of $TTD_i$ for evaluating the model on edge $i$ \\ \hline
$D$ & theoretical dataset if all of the client data were aggregated \\ \hline
$D_i$ & Edge {i} or Client ${i}$'s data; when it refers to data of Edge ${i}$, it is a theoretical dataset if all of the client data connected to edge ${i}$ were aggregated \\ \hline
$Acc(M, TD)$ & accuracy of model M on the test set TD \\ \hline 
\end{tabular}
\label{tab:notation}
\vspace{-10pt}
\end{table}
\subsection{Problem Definition}
\label{section:problem_definition}
The goal of conventional 2-level FL and 3-level FL is to improve the performance of the global model, and minimize the loss of aggregated global model against individual data points. To formulate this:
\begin{gather}
\label{eq:main}
\textstyle \min_{GM \in \mathbb{R}^d} f(GM) %\nonumber 
\end{gather}
\begin{gather}
\label{eq:main}
\textstyle f(GM) = \sum_{c=1}^{C} \frac{|D_c|}{|D|} F_c(GM) %\nonumber 
\end{gather}
\begin{gather}
\label{eq:main}
\textstyle F_c(GM) = \frac{1}{|D_c|}\sum_{i \in D_c} f_i(GM) 
\end{gather}

\noindent where $f_i(GM) = L(x_i, y_i; GM)$. $L(x_i, y_i; GM)$ is the loss of the prediction made with the global model $GM$ on example $(x_i, y_i)$. $C$ represents the set of clients, and $D = (x, y)$ denotes the theoretical dataset obtained by aggregating client data, where $x$ is the input and $y$ the corresponding label \cite{FedClassAvg}.

The goal of personalizing FL is to minimize the average loss of each client’s personalized model, particularly demonstrating its value under non-IID settings where clients possess heterogeneous data distributions \cite{FedClassAvg}. In this paper, we aim to personalize edge models in edge-accommodated 3-level FL to enhance the performance of each edge model and minimize the loss of the personalized edge model against individual data points across the devices associated with the edge. To formulate this:
\begin{gather}
\textstyle \min_{PEAM_1, PEAM_2, ... PEAM_K}\sum_{k=1}^{K} \frac{|D_k|}{|D|} F_k(PEAM_k)%, \nonumber\\ 
\label{eq:targetproblem} \end{gather}
\begin{gather}
\textstyle F_k(PEAM_k) = \frac{1}{|D_k|}\sum_{i \in D_k} f_i(PEAM_k) 
\end{gather}
where $PEAM_k$ is the personalized edge aggregated model of $k$-th edge, $f_i(PEAM_k) = L(x_i, y_i; PEAM_k)$, and $L(x_i, y_i; PEAM_k)$ is the loss of the prediction made with the personalized edge aggregated model for edge $k$, $PEAM_k$, on the example $(x_i, y_i)$ that belongs to $D_k$. Note that $D_k$ is the theoretical dataset if all of the clients' data connected to edge $k$ were aggregated.

In other words, our research focuses on scenarios where each \textit{edge has its own optimization goal and primarily aims to perform well on its unique data distribution}. Regardless of the edge-level non-IIDness--whether distributions are similar, distinct, or partially overlapping with certain labels dominating specific edges--the goal of personalizing 3-level FL is to benefit all edges by addressing the unique label distributions at each edge, as determined by their locations. This setting is motivated by real-world applications, where edges are situated in different geographic or operational contexts, encounter diverse data distributions, and where having separate personalized models per edge offers greater benefit to the associated clients than a single global model shared across edges, as further discussed in Section \ref{sec:Scenario-specifications}.

Our setting departs from conventional work on personalized federated learning, where personalization typically occurs at the client level after training a global model. Our framework eliminates client-side personalization entirely and instead performs exclusive personalization at the edge level. This distinction reflects a fundamental shift in the problem formulation: rather than each client pursuing its own optimization goal, we consider each edge node as the primary unit of personalization, with its own objective and associated personalized model. Each edge model is optimized and evaluated against a dedicated edge-level test set, representing the data distribution of its associated clients.
This formulation introduces a new challenge: hierarchical non-IIDness. Unlike traditional FL settings \textit{that focus on a single level of non-IIDness (across clients), our framework incorporates two levels}. The primary focus is on varying edge-level non-IIDness while the underlying client-level non-IIDness is assumed to be severe. To properly evaluate performance under this setting, we design test sets at the edge level, rather than at the cloud or client level, as discussed further in the following section.

In summary, this formulation redefines the personalization objective in hierarchical federated learning by shifting the focus from clients to edge nodes with distinct optimization goals, under a two-tiered non-IID structure. This motivates our approach to personalization at the edge level, along with edge-level test sets, which we detail in the following section.

\subsection{Locating Test Set at the Edge}
\label{sec:Locating-Test-Set-At-The-Edge}
To evaluate whether we are solving the problem defined in Section \ref{section:problem_definition}, it was important to design the test set correctly. As part of it, we distributed the test set across each edge, rather than centralizing it on a global server, as shown in Figure \ref{fig:PHE-FL} represented as diamonds, and ensure that each set contains all labels present in its corresponding edge training set. 

The primary reason for allocating the test set to the edge is due to a shift in goal: we are no longer concerned with the model's performance across the entire dataset; instead, we focus on the real world scenarios where each edge has its own optimization goal as defined in Equation \eqref{eq:targetproblem}. By dedicating a test set to each edge, we can precisely assess the model's performance relative to the specific data distribution at each edge. This approach distinguishes our research from studies that utilized a 3-level FL architecture without personalization, where the test set is typically maintained at the cloud level.

Locating test set at the edge aligns with the fundamental principles of FL. Traditional 2-level FL used a test set on the cloud server containing all labels to measure the loss of the aggregated global model against individual data points. This setup assumes that the connecting entity, cloud server, can maintain a test set that reflects the overall data distribution across devices, while individual devices and the cloud server lack insight into the specific data distributions of individual devices. Hence, testing the model against this test set can be a good indicator of whether the model is achieving the goal, i.e., performing well against individual data points of all devices in the network. In our research, the objective has shifted to minimizing the loss of models against the data points specific to each edge. According to the assumption made in the original 2-FL research, each edge, as a connecting entity, can maintain a test set that reflects the distribution of connected devices without knowing the specific distributions of individual devices. Hence, testing the model against the test sets located on the edges is a reliable indicator of the model's performance on individual data points specific to each edge, ensuring high edge-wise performance.

\subsection{Scenario Specifications}
\label{sec:Scenario-specifications}
In our research, we focus on real world applications where location proximity significantly influences the label distribution at the edge. The edges will have distinct data distributions, each even having different subsets of labels, resulting in different optimization goals for each edge. To better reflect real-world scenarios with varying location differences, all of our experiments are designed such that each edge has its own subset of labels and varying distributions of samples per label.

One example of the applications is in the domain of smart traffic monitoring, which has been adopted by cities such as Barcelona, London, and San Francisco, and is offered as a product by various companies \cite{smartcity,smartcity_sanfrancisco,smartcity2,smartcity_visoai}. In this context, the detection and identification of traffic patterns differ significantly based on location: school zones and parks commonly feature bicycles, pedestrians, and animals but rarely cars, while intersections have more cars and trucks but fewer pedestrians and animals. Highways are dominated by cars and trucks, with few pedestrians or animals. As the edges in different locations have different optimization goals, the personalized model should be tailored to the specific labels and data distributions relevant to each edge to ensure it meets the unique requirements and performs optimally in each context.

The cases we are focusing on represent a variety of real-world challenges. By providing personalized models for each edge, PHE-FL enhances image classification tasks in these challenges by inherently leveraging location-based contextual information. Simpler tasks, such as image classification under varying lighting conditions at different photo stations, can benefit from our method. It is also well-suited for advanced applications, such as object detection in robotics \cite{robotics_1}, where the objects to be detected depend on the robot's location, or diagnosing diseases using medical images with ML models, where disease prevalence varies based on geographic factors.

%% file: PHEFL.tex
\section{Personalized Hierarchical Edge-enabled Federated Learning}
\subsection{Motivation}
\label{sec:3-Motivation}
Two extreme cases of edge-level non-IIDness require different FL architectures to optimize performance. In the IID case, where each edge has an equal distribution of all labels, a 3-level FL EdgeCloud approach is ideal, as aggregating knowledge from all edges to create a global model allows each edge to benefit from insights across others. In contrast, in the non-IID case, when each edge has a distinct data distribution with only one label per edge, an edge-only aggregation approach (OnlyEdge) without cloud level aggregation is ideal, as global aggregation across edges provides no relevant insights to individual edges and would rather introduce noise. 

As such, we can infer that depending on the extent of edge-level non-IIDness, there are different FL architecture that leads to the highest performance, i.e., minimizing the edges' average loss on the data distributions of local devices connected to each edge. More precisely, the degree of non-IIDness determines the effectiveness of cloud-level aggregation. However, identifying this degree is challenging, as neither of edges, the cloud, and users has the complete knowledge of data distributions of each edge. Moreover, for the scenarios with intermediate levels of non-IIDness, the best architectural choice is not intuitively obvious. As far as we know,  there is no automated method exists to adjust the architecture based on non-IIDness. Thus, researchers must rely on empirical testing to select the best approach.
In response to these challenges, \textbf{our research develops a personalization method robust across all non-IID levels, automatically adjusting its parameters to achieve high performance for each edge's tasks, without prior knowledge of the data distribution across edges.}

\subsection{PHE-FL Overview}
PHE-FL is a method that personalizes a 3-level FL model at the edge, aiming to ensure robust performance despite edge-level non-IIDness. Motivated by the observation that the degree of non-IIDness determines the effectiveness of cloud-level aggregation, PHE-FL emphasizes that determining \textbf{how much knowledge each edge should adopt from other edge models} is crucial for the personalization process.

To address this, PHE-FL introduces two key advancements. First, the cloud server creates different cloud-aggregated models for each edge as shown in the equation \ref{eq:central_aggregation} and equation \ref{eq:cam_update}. This is achieved by aggregating all edge models, excluding the target edge. By evaluating a cloud-aggregated model for a specific edge, we can assess how other edges, excluding the target edge, can contribute to its performance. 
\begin{gather}
\label{eq:central_aggregation}
\textstyle |D - D_e| = \sum_{\substack{k \in K \\ k \neq e}} |D_k|
\end{gather}
\begin{gather}
\label{eq:cam_update}
\textstyle \text{CAM}_e^t = \sum_{\substack{k \in K \\ k \neq e}} \frac{|D_k|}{|D - D_e|} \, \text{EAM}_k^t
\end{gather}
Second, we personalized model on each edge following the equation \ref{eq:alpha_definition} and \ref{eq:main_personalization}. PHE-FL incorporates a dynamic parameter, $\alpha$, which weighs the extent of model adoption from its own edge compared to other edges during the personalization process. In each round, every edge calculates its $\alpha$ value, indicating how much the personalization leverages knowledge from its own edge-aggregated model, and $1-\alpha$, indicating the extent to which it leverages knowledge from other edges via the cloud-aggregated model tailored for that particular edge. The value of $\alpha$ is determined by comparing the performance of both the edge-aggregated model and the cloud-aggregated model on the edge’s own test set for personalization, $PTD_k$. This comparison helps ascertain which model contains more applicable knowledge for the specific edge and should be more heavily reflected in the personalized edge model.
\begin{gather}
\label{eq:alpha_definition}
\textstyle \alpha_k = \frac{\text{Acc}(\text{EAM}_k^t, \text{PTD}_k)}{\text{Acc}(\text{EAM}_k^t, \text{PTD}_k) + \text{Acc}(\text{CAM}_k^t, \text{PTD}_k)}
\end{gather}
\begin{gather}
\label{eq:main_personalization}
\textstyle \text{PEAM}_k^t = \alpha_k \cdot \text{EAM}_k^t + (1 - \alpha_k) \cdot \text{CAM}_k^t
\end{gather}
By creating different cloud-aggregated models and calculating unique $\alpha$ values for each edge in every round, PHE-FL develops a personalized model for each edge in every round. The details of each component are provided in the following sections.

\subsection{Cloud-Aggregated Model for Each Edge}
\textfloatsep = 0.7\baselineskip plus 0.2\baselineskip minus 0.4\baselineskip

\begin{algorithm}[b]
\footnotesize	
\begin{algorithmic}[1]
\caption{Cloud-level aggregation of PHE-FL}
\label{algo:PHE-FL-global}
    \FOR {each edge \(e \gets 1, 2, \ldots, K\)}
    \STATE \(|D - D_e| \gets \sum_{\substack{\forall k \in K, k \neq e}} \left|D_k\right| \)
    \STATE \(CAM^{t}_e \gets \sum_{\substack{\forall k \in K, k \neq e}} 
    \left( \frac{|D_k|}{|D-D_e|} \, EAM_k^{t} \right)\)
    \STATE Broadcast \(CAM^{t}_e\) to Edge \(e\) for \(\forall e \in K\)
    \ENDFOR
\end{algorithmic}
\end{algorithm}

PHE-FL constructs $|K|$ cloud-aggregated models, one for each edge, rather than a single global model, as outlined in Algorithm \ref{algo:PHE-FL-global}. For each edge $e$, its cloud-aggregated model is formed by aggregating the models of all other $|K| - 1$ edges, excluding $e$'s own model. Our personalization aims to find the $\alpha$ value that precisely assesses the relevance and helpfulness of knowledge from other edges for each specific edge. Including $e$'s own model would overestimate cloud aggregated model's performance, as it would incorporate $e$'s knowledge. Excluding its own model ensures an accurate evaluation of how other edges contribute to $e$. It is worth noting that this process does not burden the client models compared to traiditiional FL, as the extra computation occurs at the cloud level, which is assumed to have sufficient computational resources and storage.

\subsection{Personalization on Edge with Dynamic Parameter $\alpha$}

\begin{algorithm}[tb]
\footnotesize	
\begin{algorithmic}[1]
\caption{Model Mixture Personalization of PHE-FL}
\label{algo:PHE-FL-alpha}
    \FOR {each edge \( k \in K\)}
        \STATE $EdgeModelAcc$ $\gets$ $Acc(EAM_k^{t}, PTD_k)$
        \STATE $CloudModelAcc$ $\gets$ $Acc(CAM^{t}_k, PTD_k)$
        \STATE $\alpha_k$ $\gets$ \(\frac{EdgeModelAcc}{EdgeModelAcc + CloudModelAcc}\)
        \STATE \(PEAM_k^{t}\) $\gets$ $\alpha_k$ * $EAM_k^{t}$ + $(1-\alpha_k)$ * $CAM^{t}_k$
        \STATE Broadcast $PEAM_k^{t}$ for \(\forall c \in E_k\)
    \ENDFOR
\end{algorithmic}
\end{algorithm}

The second unique aspect of PHE-FL is that it mixes the cloud-aggregated and edge-aggregated models to create a personalized edge-aggregated model, as shown in Algorithm \ref{algo:PHE-FL-alpha} at the end of each round. Each edge $k$ calculates two accuracy scores on its 15\% test set $PTD_k$, drawn from $TTD_k$, using both its own model $EAM_k^t$ and the cloud-aggregated model $CAM_k^t$. The edge then computes the \textbf{dynamic} parameter $\alpha$, which reflects the contribution of the edge model relative to the cloud model. Specifically, $\alpha$ is calculated by dividing the accuracy of the edge-aggregated model by the sum of the accuracies of both models as defined in Equation \ref{eq:alpha_definition}. PHE-FL uses this $\alpha$ to linearly combine the edge-aggregated model, weighted by $\alpha$, and the cloud-aggregated model, weighted by $1-\alpha$, to form the personalized model as defined in Equation \ref{eq:main_personalization}. The method for extracting $PTD_k$ is in Section \ref{sec:data-partition}.

By excluding model $k$ from the cloud-aggregated model, the accuracy of $CAM_k^t$ against $PTD_k$ demonstrates the value of knowledge from other edges for $k$. Comparing this accuracy to that of $EAM_k^t$ shows the relative knowledge each model provides for edge $k$, and $\alpha$ represents their knowledge ratio. Weighting the models according to their accuracy refines the personalized model.

Personalization step imposes no additional burden on the client models compared to traditional FL, as the extra computation occurs at the cloud level, which is assumed to have sufficient computational power,
communication capability, and storage.

\subsection{PHE-FL Workflow}
In this section, we illustrate the complete workflow of PHE-FL as shown in Figure \ref{fig:PHE-FL} and Algorithm \ref{algorithm:PHE-FL}. The PHE-FL process begins with the cloud server initializing a global model $GM^0$ and broadcasting it to individual edge servers. For each round $t$, all IoT devices, specifically the $c$-th IoT device in the $k$-th edge, train the model $DM_c^{t-1}$. Trained model $DM_c^t$ is then sent to the $k$-th edge server. The edge server uses the FedAvg algorithm to aggregate these models and obtain $EAM_k^t$, which is then transmitted to the cloud server. The cloud server performs cloud-level aggregation and creates $|K|$ cloud-aggregated models, as shown in Algorithm \ref{algo:PHE-FL-global}. These $|K|$ models are then sent to the corresponding edges. After receiving the global model, each edge server creates the personalized model $PEAM_k^t$ using Algorithm \ref{algo:PHE-FL-alpha}. In the following ${t+1}$-th round, the $c$-th IoT device in the $k$-th edge will train the model $PEAM_k^t$, iterating through the training, aggregations, and personalization steps outlined above.

\begin{algorithm}[tb]
\setstretch{0.9}
\footnotesize	
\caption{PHE-FL : Personalized Hierarchical Edge-enabled Federated Learning}
\label{algorithm:PHE-FL}
\begin{algorithmic}[1]

\STATE INPUT : $epochs$, $batchSize$, $K$, $E_k$ for \(\forall k \in K \)

\STATE OUTPUT : $PEAM_k$ for  \(\forall k \in K\)
\STATE

\STATE $\triangleright$ Initialization
\STATE Cloud Server initializes $GM^0$
\STATE Broadcast $GM^0$ to all edges and devices

\STATE
\STATE $\triangleright$ {Iteration}
\FOR {each round $t \gets 1, 2, \ldots T$}
    \STATE $\triangleright$ {IoT devices train model}
    \FOR {each edge \( k \in K\)}
        \FOR {each IoT device \(c \in E_k\)}
            \STATE \(DM_c^{t} \gets ClientTrain(DM_c^{t-1}, epochs, batchSize) \)
            \STATE Transmit ($DM_c^{t}$, $|D_c|$) to Edge $k$
        \ENDFOR
    \ENDFOR

    \STATE
    \STATE \(\triangleright\) Edge level aggregation
    \FOR {each edge \( k \in K\)}
        \STATE \(|D_k| \gets \sum_{\substack{c \in E_k}} \left|D_c\right| \)
        \STATE \(EAM_k^{t} \gets 
       \sum_{\substack{c \in E_k}} \left( \frac{|D_c|}{|D_k|} DM_c^{t} \right)\)
        \STATE Transmit ($EAM_k^{t}$, $|D_k|$) to Cloud Server
    \ENDFOR

    \STATE
    \STATE \textbf{Cloud level aggregation} by Algorithm \ref{algo:PHE-FL-global}
    \STATE \textbf{Model Mixture Personalization} by Algorithm \ref{algo:PHE-FL-alpha}
    \STATE
    \STATE \( DM_c^{t} \gets PEAM_k^{t} \) \textbf{for} \( \forall c \in E_k, \forall k \in K \)

\ENDFOR
\end{algorithmic}
\end{algorithm}

%% file: experiment.tex
\section{Experiments}
\subsection{Experimental Settings}
\subsubsection{Data Set}
For our image classification experiment, we
used MNIST \cite{MNIST}, F-MNIST \cite{Fashioin_MNIST} and CIFAR10 \cite{CIFAR10} datasets.
\subsubsection{Model Architecture and Hyperparameters}
We used the CNN architecture from FedAvg \cite{mcmahan2017}. The input layer accepts 28x28x1 images for MNIST/F-MNIST and 32x32x3 images for CIFAR10. The network includes two Conv2D layers (32 and 64 filters, 5x5 kernels) with ReLU activation and ‘same’ padding, followed by 2x2 MaxPooling2D. After flattening, we added 2 Dense layers: one with 512 units (ReLU) and another with 10 units (softmax). This CNN has 1,663,370 parameters for MNIST/F-MNIST and 2,156,490 for CIFAR10.

Each client’s CNN model was trained with the SGD optimizer (learning rate 0.1) for 5 epochs per FL round, using a batch size of 32 and sparse categorical cross-entropy loss with accuracy as the metric.

Throughout our results section, we define a `round' as the process starting from client devices initiating training, proceeding to edge and global level aggregation, broadcasting from the cloud server to the edge, personalization if applicable, and concluding with broadcasting from the edge to the client.

Our network includes 1 global server, 10 edges, and 100 IoT devices (10 devices per edge), using \texttt{c5.24xlarge} and \texttt{m6a.24xlarge} instances with \texttt{Linux/UNIX} OS.

\subsubsection{Baselines}
\begin{figure}[tb]
\centering
\includegraphics[width=8cm]{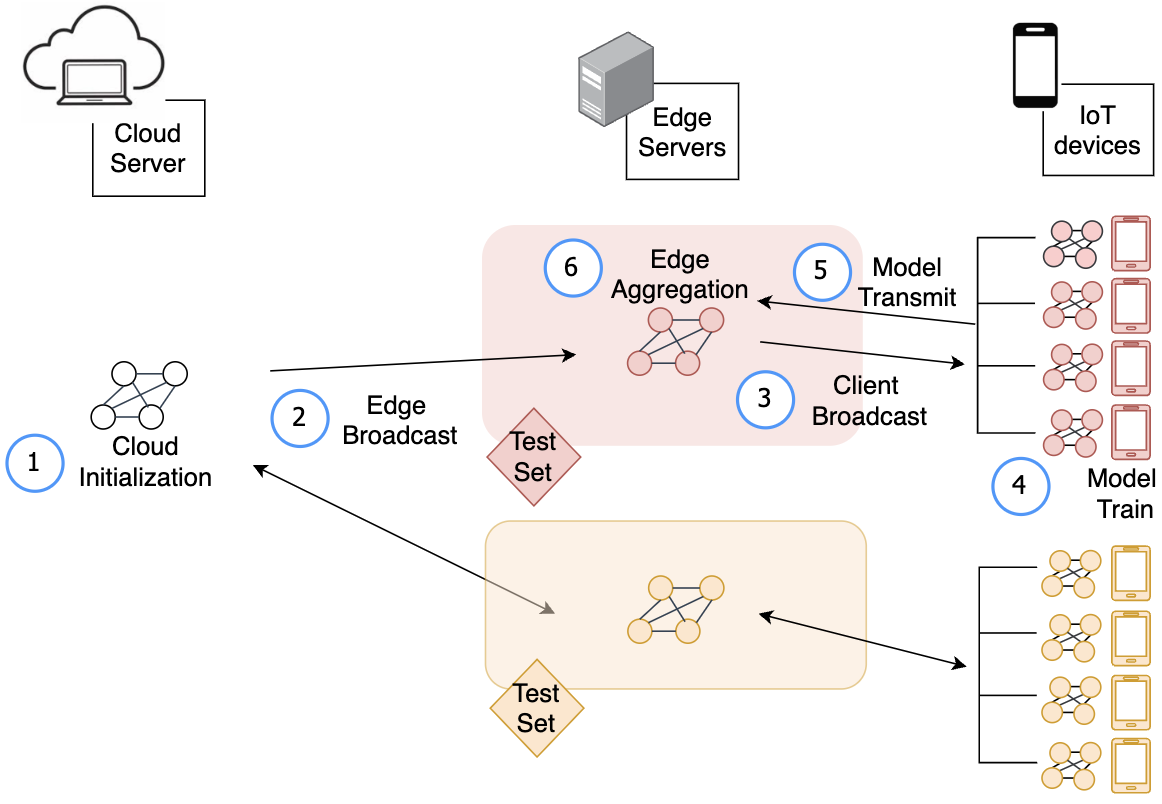}
\caption{OnlyEdge Workflow: (1) Cloud server initializes the global model for the first round. (2) Global model is distributed to the edge servers. (3) Each edge server either distributes the initial global model in the first round or the aggregated edge model in subsequent rounds. (4) IoT devices train the edge model with local data. (5) IoT devices send the model weights back to the edge server. (6) Edge servers aggregate the models from IoT devices. Repeat (3) to (6).}
\label{fig:OnlyEdge}
\setlength{\belowcaptionskip}{-30pt}
\end{figure}

To thoroughly evaluate the performance of PHE-FL, we compared it to two baselines:
\begin{itemize}
  \item[\textbullet] \textbf{EdgeCloud}: EdgeCloud is the current state-of-art of 3-level FL. Individual device models are first aggregated at the edge, edge aggregated models are then aggregated at the cloud, then the global model is uniformly distributed to all clients.
  
  \item[\textbullet] \textbf{OnlyEdge}: OnlyEdge is a method that aggregates weights only at the edge level, without involving a cloud server, as shown in the Figure \ref{fig:OnlyEdge} and equations \ref{eq:Only_Edge}. 
  \begin{gather}
    \label{eq:Only_Edge}
    \textstyle \text{EAM}_k^t = \sum_{c \in E_k} \frac{|D_c|}{|D_k|} \, \text{DM}_c^t
  \end{gather}
  The Edge Aggregated Model (\(\text{EAM}_k^t\)) of edge \(k\) is the aggregated model of the individual device models (\(\text{DM}_c^t\)) belonging to edge \(k\).
  
This approach mirrors traditional FL by distributing aggregation across edge servers instead of a central cloud. As this architecture is inherent in traditional FL, it has not been explicitly addressed in existing research. We define it to evaluate and compare the performance of traditional FL models at the edge level.
\end{itemize}

\subsubsection{Evaluation Metrics}
We used two primary evaluation metrics to compare the performance of the three different models.

The first metric, \textbf{Acc}$\boldsymbol{N}$, represents the highest accuracy achieved within $N$ rounds. After each training round, we evaluated the global model (EdgeCloud), the edge-aggregated models (OnlyEdge), and the personalized edge models (PHE-FL) against 10 edge test sets for evaluation, $ETD_k$, averaging the 10 accuracy scores to obtain the Acc$N$. 

We have employed a second metric, \textbf{Drop}$\boldsymbol{M}$, to assess performance fluctuation and stability under non-IID data. DropM represents the maximum accuracy drop after the model reaches M\% accuracy. Once this threshold is met, we define a 10-round window and calculate the difference between the maximum and minimum accuracy, recording the largest value. A lower DropM indicates improved stability, as non-IID data can cause sudden accuracy drops. If the model doesn't reach M\%, DropM cannot be measured.

\subsection{Data Partitions}
\label{sec:data-partition}
\subsubsection{Train Data Partitions}
In our research, we explored edge-level non-IID data influenced by the number of labels per edge and the number of samples per label for each edge. Specifically, we considered non-IIDness due to \textbf{skewed label distribution at the edge level}. To achieve this, we assumed \textbf{the most extreme case of non-IIDness at the device level}, constrained by a given edge-level non-IIDness. In this setup, each IoT device has only one label in its training set. For a classification task with 10 labels, this means that only 10 devices among the 100 devices share the same label. Moreover, we ensured that each device holds a similar volume of data, and that every edge connects to an equal number of devices, resulting in hypothetical training datasets at the edge level being of equal size. This setup allows us to clearly identify the impact of data heterogeneity caused by the number of labels and the number of samples per label on 3-level FL.

The basic intuition behind the effects of the number of labels per edge and the sample distribution per label on edge-level non-IID is straightforward: \textbf{increasing the number of labels per edge tend to make the edge-level data distribution more IID}, as more labels are likely to overlap across edges. Conversely, \textbf{increasing the sample count for individual labels tends to enhance edge-level non-IIDness}. This is because when an edge predominantly features one label, considering the limited number of samples per label for the entire dataset, fewer edges will contain this label, resulting in increased heterogeneity across the edges.

\begin{table}[tb]
\caption{Overview of 4 Train Set Distribution Scenarios}
\label{tab:six-scenarios}
\centering
\scriptsize
\begin{tabular}{| p{2.0cm} | p{1.9cm} | p{3.4cm} |}
\hline
\textbf{Scenario Name} & \textbf{Number of Labels per Edge} & \textbf{Predominant Label Proportion} \\ \hline
Distribution 1 (D1)& 1 & 100\% of the edge's data \\ \hline
Distribution 2 (D2)& 5 & 20\% of the edge's data \\ \hline
Distribution 3 (D3)& 8 & 30\% of the edge's data \\ \hline
Distribution 4 (D4)& 10 & 10\% of the edge's data \\ \hline
\end{tabular}
\vspace{-10pt}
\end{table}

Our study includes 4 scenarios with different edge-level non-IIDness to address the above factors. These 4 scenarios are summarized in Table \ref{tab:six-scenarios}. These scenarios cover 4 different numbers of labels per edge: 1, 5, 8, and 10. The maximum number of devices within an edge that have the same label * 10\% is indicated as the Predominant label proportion. In Distribution 1, all 10 devices within an edge share the same label. In Distribution 2, 5 pairs of devices each share a label. In Distribution 3, 3 devices share one label, and the remaining 7 each have different labels (see Table \ref{table:ExampleImbalancedTest}). In Distribution 4, all 10 devices within an edge have different labels. 

Edge-level IIDness progressively increases from D1 to D4. Specifically, in D1, edges do not share any labels, representing an extreme non-IID scenario. Moving sequentially through  D2 and D3, the number of labels shared per edge gradually increases, while the dominance of specific labels diminishes. This transition strengthens edge-level IIDness. By D4, all labels are shared across edges, representing an extreme IID setting.

\subsubsection{Balanced and Imbalanced Test Set ($TTD$) }

\begin{table*}[!hb]
\caption{Label Breakdown of Training Set and Imbalanced Test Set ($TTD$) for Distribution 3}
\label{table:ExampleImbalancedTest}
\centering
\scriptsize 
\begin{tabular}{|c|c|c|c|c|c|c|c|c|c|c|c|}
  \hline
  (unit: \%) & label0 & label1 & label2 & label3 & label4 & label5 & label6 & label7 & label8 & label9 & TOTAL \\ \hline 
  Edge 1 & 30 & 10 & 10 & 10 & 10 & 10 & 10 & 10 & 0 & 0 & 100 \\ \hline
  Edge 2 & 0 & 30 & 10 & 10 & 10 & 10 & 10 & 10 & 10 & 0 & 100 \\ \hline
  Edge 3 & 0 & 0 & 30 & 10 & 10 & 10 & 10 & 10 & 10 & 10 & 100 \\ \hline
  Edge 4 & 10 & 0 & 0 & 30 & 10 & 10 & 10 & 10 & 10 & 10 & 100 \\ \hline
  Edge 5 & 10 & 10 & 0 & 0 & 30 & 10 & 10 & 10 & 10 & 10 & 100 \\ \hline
  Edge 6 & 10 & 10 & 10 & 0 & 0 & 30 & 10 & 10 & 10 & 10 & 100 \\ \hline
  Edge 7 & 10 & 10 & 10 & 10 & 0 & 0 & 30 & 10 & 10 & 10 & 100 \\ \hline
  Edge 8 & 10 & 10 & 10 & 10 & 10 & 0 & 0 & 30 & 10 & 10 & 100 \\ \hline
  Edge 9 & 10 & 10 & 10 & 10 & 10 & 10 & 0 & 0 & 30 & 10 & 100 \\ \hline
  Edge 10 & 10 & 10 & 10 & 10 & 10 & 10 & 10 & 0 & 0 & 30 & 100 \\ \hline
\end{tabular}
\end{table*}

\begin{table*}[!h]
\caption{Label Breakdown of Balanced Test Set ($TTD$) for Distribution 3}
\label{table:ExampleBalancedTest}
\centering
\scriptsize 
\begin{tabular}{|c|c|c|c|c|c|c|c|c|c|c|c|}
  \hline
  (unit: \%) & label0 & label1 & label2 & label3 & label4 & label5 & label6 & label7 & label8 & label9 & TOTAL \\ \hline 
  Edge 1 & 100 & 100 & 100 & 100 & 100 & 100 & 100 & 100 & 0 & 0 & 800 \\ \hline
  Edge 2 & 0 & 100 & 100 & 100 & 100 & 100 & 100 & 100 & 100 & 0 & 800 \\ \hline
  Edge 3 & 0 & 0 & 100 & 100 & 100 & 100 & 100 & 100 & 100 & 100 & 800 \\ \hline
  Edge 4 & 100 & 0 & 0 & 100 & 100 & 100 & 100 & 100 & 100 & 100 & 800 \\ \hline
  Edge 5 & 100 & 100 & 0 & 0 & 100 & 100 & 100 & 100 & 100 & 100 & 800 \\ \hline
  Edge 6 & 100 & 100 & 100 & 0 & 0 & 100 & 100 & 100 & 100 & 100 & 800 \\ \hline
  Edge 7 & 100 & 100 & 100 & 100 & 0 & 0 & 100 & 100 & 100 & 100 & 800 \\ \hline
  Edge 8 & 100 & 100 & 100 & 100 & 100 & 0 & 0 & 100 & 100 & 100 & 800 \\ \hline
  Edge 9 & 100 & 100 & 100 & 100 & 100 & 100 & 0 & 0 & 100 & 100 & 800 \\ \hline
  Edge 10 & 100 & 100 & 100 & 100 & 100 & 100 & 100 & 0 & 0 & 100 & 800 \\ \hline
\end{tabular}
\end{table*}

In our research, we used 2 distinct types of test sets for each training set distribution. The `imbalanced' test set mirrors the data distribution of the training set. In contrast, the `balanced' test set contains an equal proportion of samples for each label associated with the edge, regardless of their proportions in the training set. To create the balanced test set, we included all the available samples for each label. As the entire test set contains a similar number of samples per label, the resulting test set has a `balanced' proportion of data across labels. 

For example, in Distribution 3, the imbalanced test set, as shown in Table \ref{table:ExampleImbalancedTest}, ensures that each edge's test set is proportional to its corresponding training set's data distribution. Conversely, the balanced test set for the same Distribution 3 includes an entire test set of each label that edge has, ensuring comprehensive coverage. As shown in Table \ref{table:ExampleBalancedTest}, each edge will have all of the test sets of labels that are associated with the train set. Since we are considering datasets with an equal number of samples per label in the test set, the balanced set is larger by a factor equal to the number of labels per edge. For example, in Distributions 3, the balanced test set is 8 times larger than the imbalanced one.

Note that \textbf{the implications of both balanced and imbalanced test sets for Distribution 1, Distribution 2, and Distribution 4 are the same, as these scenarios have an equal proportion of samples per label for each edge}. For these scenarios, we have included all samples in the test set that have corresponding labels in the edge's training set. We labeled them as `imbalanced' test sets since these test sets are proportional to the training data anyway -- these represent the `worst imbalanced test set' that each scenario could get.

The rationale for having two test sets for each case stems from the assumptions of FL introduced in the Section \ref{sec:Locating-Test-Set-At-The-Edge}, that the edge, as a connectivity entity, can have a test set that reflects the overall data distribution of associated devices. This justifies the edge having an imbalanced test set that accurately represents the data distribution of the devices' train set associated with the edge. The approach of using an imbalanced test set that mirrors the imbalanced nature of the training set is widely adopted in many studies \cite{ImbalancedTest,ImbalancedTest-2}. While some studies do not explicitly address the balancedness between the training and test sets, the traditional practice of randomly splitting the dataset into training and testing implies that the imbalancedness in the dataset will carry over to both sets. Likewise, utilizing an imbalanced test set is a valid approach.

However, based on the assumption in FL that devices at the edge will not encounter labels that are not already present in any device’s current training set, but may later receive labels associated with other devices sharing the same edge, and considering the applications of this 3-level FL, where each edge device handles data with the same set of labels but with dynamic data distribution, it is also justified to use a balanced test set. This balanced test set represents all the possible data that each device could encounter in the future.

By experimenting with both test sets, we aim to thoroughly assess the robustness of our method under different potential non-IID environments that may arise from the same training dataset distribution.

\subsubsection{Test Set Partitioning for Personalization}
From the test set $TTD_i$ assigned to edge-$i$, we randomly selected 15\% to use for personalization $PTD_i$, and allocated the remaining 85\% for evaluating accuracy, $ETD_i$. For EdgeCloud and OnlyEdge evaluations, each edge also used the same $ETD_i$. 

%% file: result.tex
\section{Results}

\begin{figure*}[htbp]
\centering

  \begin{subfigure}{.03\textwidth}
  \end{subfigure}%
  \begin{subfigure}{.19\textwidth}
    \centering
    \moveright 20pt \hbox{\text{D1 imbalanced}}
  \end{subfigure}%
  \begin{subfigure}{.19\textwidth}
    \centering
    \moveright 40pt \hbox{\text{D2 imbalanced}}
  \end{subfigure}%
  \begin{subfigure}{.19\textwidth}
    \centering
    \moveright 40pt \hbox{\text{D3 imbalanced}}
  \end{subfigure}%
  \begin{subfigure}{.19\textwidth}
    \centering
    \moveright 40pt \hbox{\text{D3 balanced}}
  \end{subfigure}%
  \begin{subfigure}{.19\textwidth}
    \centering
    \moveright 40pt \hbox{\text{D4 imbalanced}}
  \end{subfigure}

\vspace{5pt}
  \begin{subfigure}{.03\textwidth}
    \centering
    \rotatebox[origin=c]{90}{\raisebox{3.5cm}{\text{MNIST}}}
  \end{subfigure}%
  \begin{subfigure}{.19\textwidth}
    \centering
    \includegraphics[width=\linewidth]{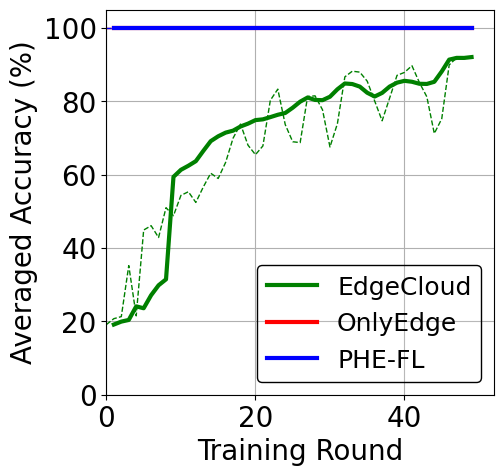}
  \end{subfigure}%
  \begin{subfigure}{.19\textwidth}
    \centering
    \includegraphics[width=\linewidth]{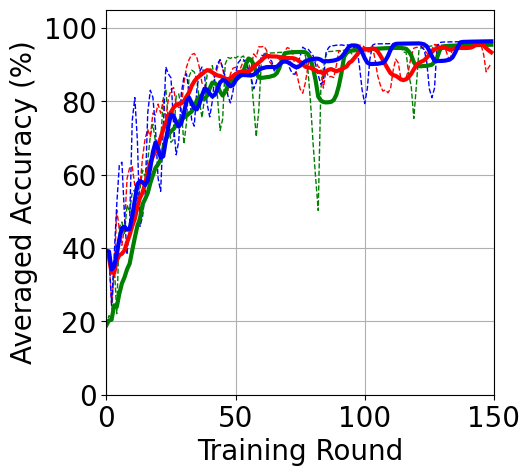}
  \end{subfigure}%
  \begin{subfigure}{.19\textwidth}
    \centering
    \includegraphics[width=\linewidth]{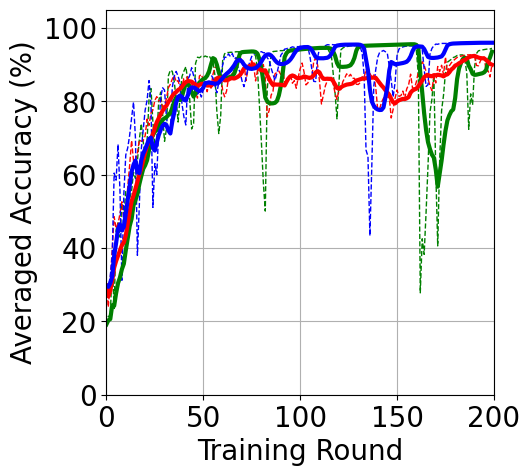}
  \end{subfigure}%
  \begin{subfigure}{.19\textwidth}
    \centering
    \includegraphics[width=\linewidth]{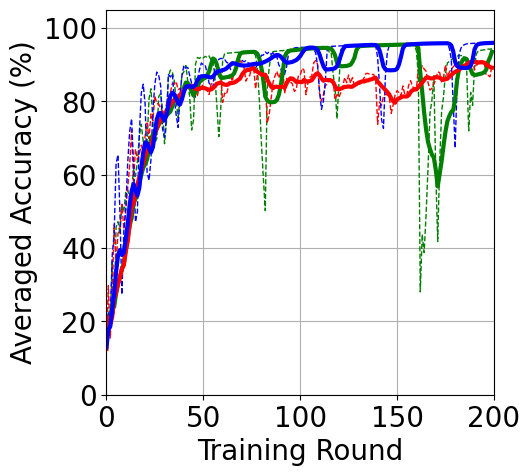}
  \end{subfigure}%
  \begin{subfigure}{.19\textwidth}
    \centering
    \includegraphics[width=\linewidth]{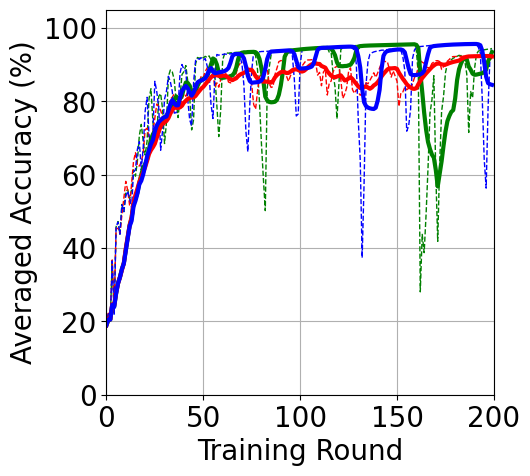}
  \end{subfigure}

  \begin{subfigure}{.03\textwidth}
    \centering
    \rotatebox[origin=c]{90}{\raisebox{3.5cm}{\text{F-MNIST}}}
  \end{subfigure}%
  \begin{subfigure}{.19\textwidth}
    \centering
    \includegraphics[width=\linewidth]{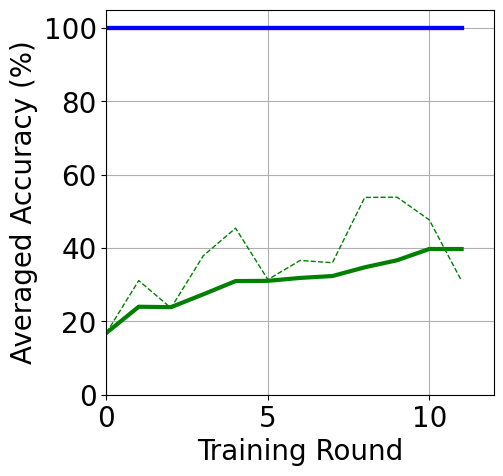}
  \end{subfigure}%
  \begin{subfigure}{.19\textwidth}
    \centering
    \includegraphics[width=\linewidth]{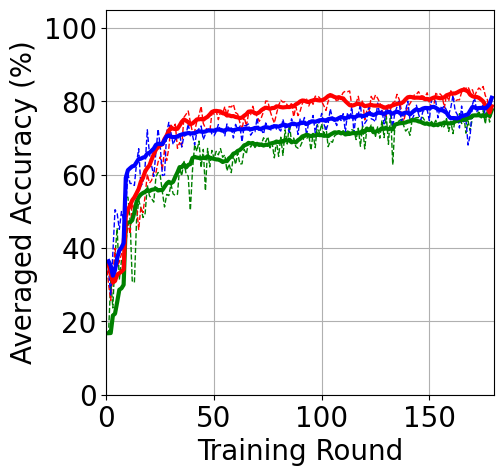}
  \end{subfigure}%
  \begin{subfigure}{.19\textwidth}
    \centering
    \includegraphics[width=\linewidth]{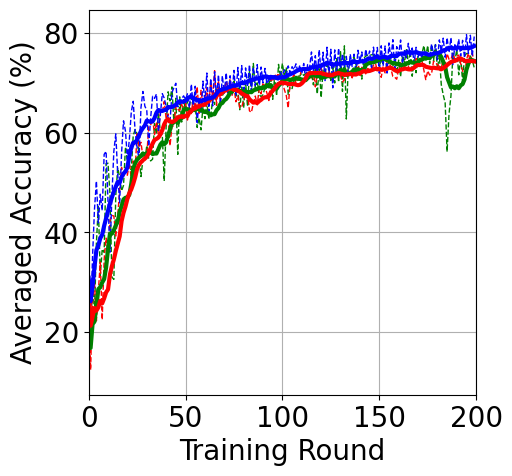}
  \end{subfigure}%
  \begin{subfigure}{.19\textwidth}
    \centering
    \includegraphics[width=\linewidth]{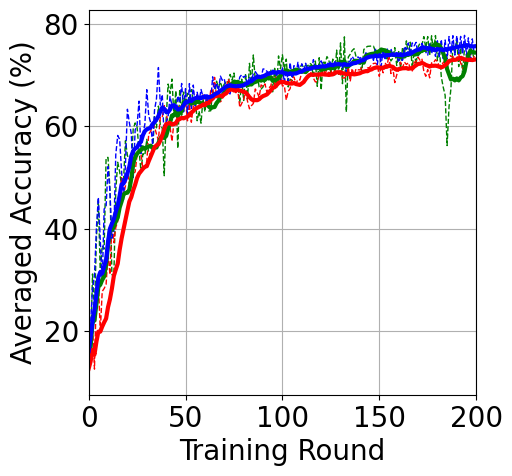}
  \end{subfigure}%
  \begin{subfigure}{.19\textwidth}
    \centering
    \includegraphics[width=\linewidth]{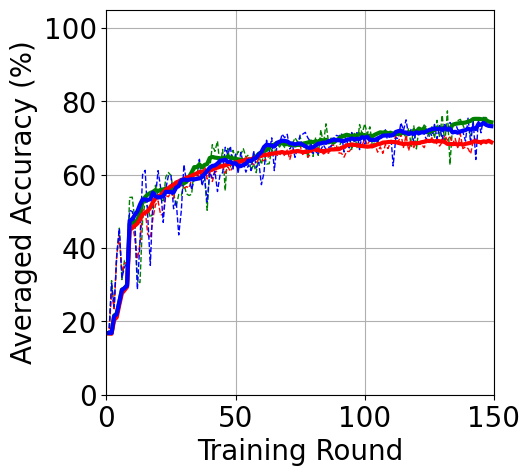}
  \end{subfigure}

  \begin{subfigure}{.03\textwidth}
    \centering
    \rotatebox[origin=c]{90}{\raisebox{3.5cm}{\text{CIFAR10}}}
  \end{subfigure}%
  \begin{subfigure}{.19\textwidth}
    \centering
    \includegraphics[width=\linewidth]{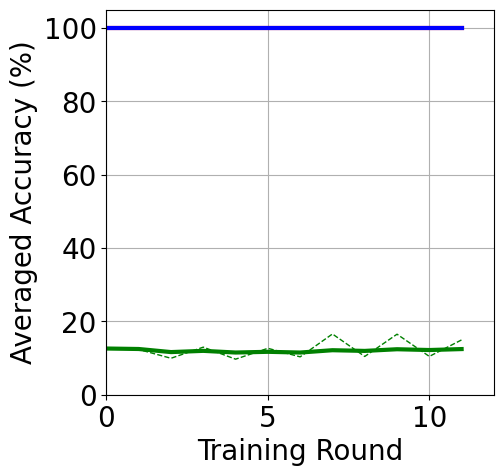}
  \end{subfigure}%
  \begin{subfigure}{.19\textwidth}
    \centering
    \includegraphics[width=\linewidth]{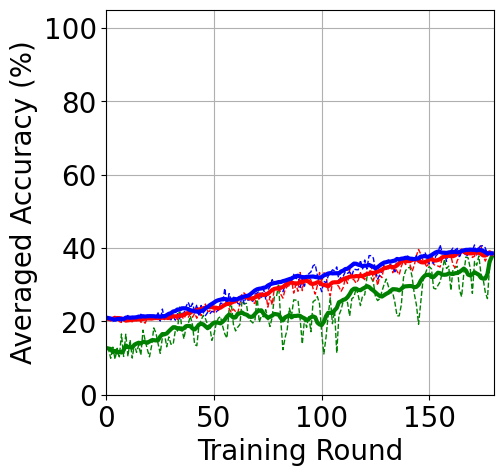}
  \end{subfigure}%
  \begin{subfigure}{.19\textwidth}
    \centering
    \includegraphics[width=\linewidth]{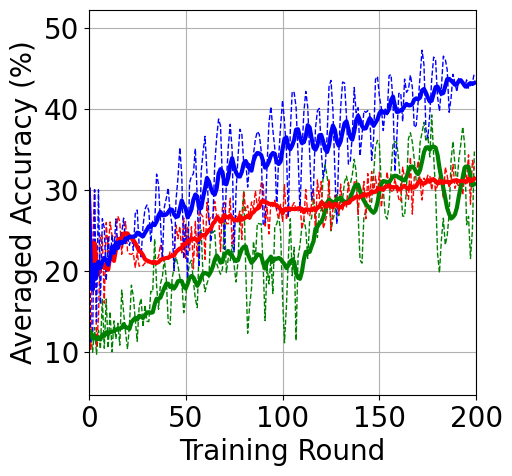}
  \end{subfigure}%
  \begin{subfigure}{.19\textwidth}
    \centering
    \includegraphics[width=\linewidth]{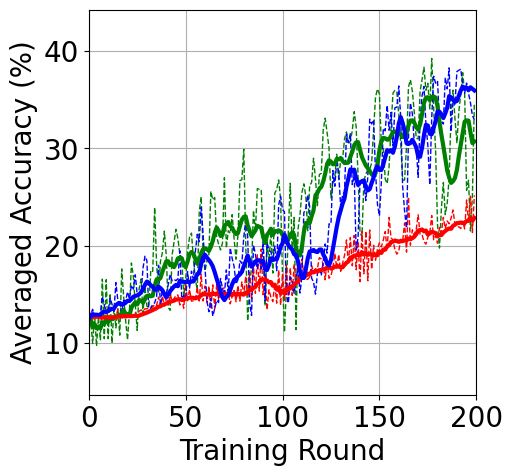}
  \end{subfigure}%
  \begin{subfigure}{.19\textwidth}
    \centering
    \includegraphics[width=\linewidth]{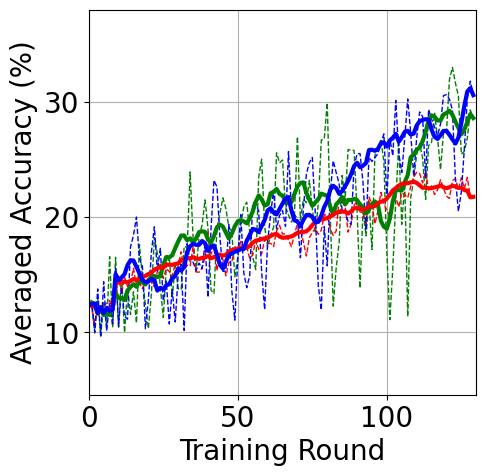}
  \end{subfigure}

  \caption{Performance of EdgeCloud, OnlyEdge, and PHE-FL on datasets (MNIST, F-MNIST, CIFAR10) across distributions D1–D6. EdgeCloud had the lowest accuracy on D1 and D2, OnlyEdge on D5 and D6, while PHE-FL never recorded the lowest accuracy.}
  \label{fig:distribution_comparison}
\end{figure*}

\begin{table*}
\caption{Results for \textbf{MNIST}, \textbf{F-MNIST}, and \textbf{CIFAR10} for different train set distributions and test sets}
\centering
\scriptsize

\setlength{\tabcolsep}{2pt} 
\renewcommand{\arraystretch}{1.2} 
\begin{tabular}{p{1cm} p{1.1cm} p{0.8cm} p{0.8cm} p{0.8cm} p{0.8cm} p{0.8cm} p{0.8cm} p{0.8cm} p{0.8cm} p{0.8cm} p{0.8cm} p{0.8cm} p{0.8cm} p{0.8cm} p{0.8cm} p{0.8cm} p{0.8cm} p{0.8cm} p{0.8cm}}
\hline
\multicolumn{1}{l}{} & \multicolumn{1}{l}{} & \multicolumn{2}{l}{D1} & \multicolumn{2}{l}{D2} & \multicolumn{2}{l}{D3} &
\multicolumn{2}{l}{D3} &
\multicolumn{2}{l}{D4} 
\\ \hline

 & \multicolumn{1}{c}{} & \multicolumn{2}{l}{Imbalanced} & \multicolumn{2}{l}{Imbalanced} & \multicolumn{2}{l}{Imbalanced} & \multicolumn{2}{l}{Balanced} & \multicolumn{2}{l}{Imbalanced} \\ \hline

%%%%%%%%%%%%%%%%%%%%%%MNIST%%%%%%%%%%%%%%%%%%%%%%%%%%%%%%%%%

\multirow{ 4}{*}{MNIST} & \multicolumn{1}{c}{} & \multicolumn{1}{c}{Acc50} & \multicolumn{1}{c}{Drop0} & 
\multicolumn{1}{c}{Acc150} & \multicolumn{1}{c}{Drop90} & 
\multicolumn{1}{c}{Acc200} & \multicolumn{1}{c}{Drop90} &
\multicolumn{1}{c}{Acc200} & \multicolumn{1}{c}{Drop90} & 
\multicolumn{1}{c}{Acc200} & \multicolumn{1}{c}{Drop90}  \\ 

& EdgeCloud              & \multicolumn{1}{c}{92.08} & \multicolumn{1}{c}{35.14} & \multicolumn{1}{c}{95.54} & \multicolumn{1}{c}{43.59} & 
\multicolumn{1}{c}{95.77} & \multicolumn{1}{c}{67.99} & 
\multicolumn{1}{c}{95.75} & \multicolumn{1}{c}{67.68} & 
\multicolumn{1}{c}{95.75} & \multicolumn{1}{c}{67.69} \\

& OnlyEdge               & \multicolumn{1}{c}{100}   & \multicolumn{1}{c}{0}    & \multicolumn{1}{c}{95.59} & \multicolumn{1}{c}{12.59} & 
\multicolumn{1}{c}{92.72} & \multicolumn{1}{c}{15.31} & 
\multicolumn{1}{c}{91.37} & \multicolumn{1}{c}{15.66} & 
\multicolumn{1}{c}{92.42} & \multicolumn{1}{c}{12.9} \\ 

& PHE-FL                 & \multicolumn{1}{c}{100}   & \multicolumn{1}{c}{0}    & \multicolumn{1}{c}{96.45} & \multicolumn{1}{c}{16.32} &
\multicolumn{1}{c}{96.1} & \multicolumn{1}{c}{52.16} &
\multicolumn{1}{c}{96.01} & \multicolumn{1}{c}{28.67} & 
\multicolumn{1}{c}{95.7} & \multicolumn{1}{c}{57.65} \\ \hline

%%%%%%%%%%%%%%%%%%%%%%FASHION%%%%%%%%%%%%%%%%%%%%%%%%%%%%%%%%%

\multirow{4}{*}{F-MNIST} & \multicolumn{1}{c}{} & \multicolumn{1}{c}{Acc12} & \multicolumn{1}{c}{Drop0} &   
\multicolumn{1}{c}{Acc180} & \multicolumn{1}{c}{Drop75} &
\multicolumn{1}{c}{Acc200} & \multicolumn{1}{c}{Drop70} &
\multicolumn{1}{c}{Acc200} & \multicolumn{1}{c}{Drop70} &
\multicolumn{1}{c}{Acc150} & \multicolumn{1}{c}{Drop75} \\ 

& EdgeCloud              & \multicolumn{1}{c}{53.88} & \multicolumn{1}{c}{37.02} & 
\multicolumn{1}{c}{77.67} & \multicolumn{1}{c}{14.67} & 
\multicolumn{1}{c}{77.67} & \multicolumn{1}{c}{21.47} & 
\multicolumn{1}{c}{77.67} & \multicolumn{1}{c}{21.47} &
\multicolumn{1}{c}{77.41} & \multicolumn{1}{c}{14.67} \\ 

& OnlyEdge               & \multicolumn{1}{c}{100}   & \multicolumn{1}{c}{0}    &  
\multicolumn{1}{c}{84.06} & \multicolumn{1}{c}{8.79} & 
\multicolumn{1}{c}{75.94} & \multicolumn{1}{c}{8.09} & 
\multicolumn{1}{c}{74.27} & \multicolumn{1}{c}{5.75} & 
\multicolumn{1}{c}{71.21} & \multicolumn{1}{c}{-} \\ 

& PHE-FL                 & \multicolumn{1}{c}{100}   & \multicolumn{1}{c}{0}    &  \multicolumn{1}{c}{81.65} & \multicolumn{1}{c}{13.57} & 
\multicolumn{1}{c}{79.66} & \multicolumn{1}{c}{8.15} & 
\multicolumn{1}{c}{77.72} & \multicolumn{1}{c}{13.87} & 
 \multicolumn{1}{c}{75.75} & \multicolumn{1}{c}{11.61} \\ \hline

%%%%%%%%%%%%%%%%%%%%%%CIFAR10%%%%%%%%%%%%%%%%%%%%%%%%%%%%%%%%%
\multirow{ 4}{*}{CIFAR10} & \multicolumn{1}{c}{} & \multicolumn{1}{c}{Acc12} & \multicolumn{1}{c}{Drop0} & 
\multicolumn{1}{c}{Acc180} & \multicolumn{1}{c}{Drop30} & 
\multicolumn{1}{c}{Acc200} & \multicolumn{1}{c}{Drop20} &
\multicolumn{1}{c}{Acc200} & \multicolumn{1}{c}{Drop20} & 
\multicolumn{1}{c}{Acc130} & \multicolumn{1}{c}{Drop0} \\ 

& EdgeCloud              & 
\multicolumn{1}{c}{16.56} & \multicolumn{1}{c}{6.87} & 
\multicolumn{1}{c}{37.85} & \multicolumn{1}{c}{15.71} & 
\multicolumn{1}{c}{39.23} & \multicolumn{1}{c}{19.48} & 
\multicolumn{1}{c}{39.23} & \multicolumn{1}{c}{19.48} & 
\multicolumn{1}{c}{32.99} & \multicolumn{1}{c}{17.64} \\ 

& OnlyEdge               & 
\multicolumn{1}{c}{100}   & \multicolumn{1}{c}{0} & 
\multicolumn{1}{c}{40.56} & \multicolumn{1}{c}{7.23} & 
\multicolumn{1}{c}{34.31} & \multicolumn{1}{c}{7.01} & 
\multicolumn{1}{c}{25.16} & \multicolumn{1}{c}{5.88} & 
\multicolumn{1}{c}{23.95} & \multicolumn{1}{c}{4.07} \\ 

& PHE-FL                 & 
\multicolumn{1}{c}{100}   & \multicolumn{1}{c}{0}    & 
\multicolumn{1}{c}{40.89}  & \multicolumn{1}{c}{7.45} & 
\multicolumn{1}{c}{47.22}  & \multicolumn{1}{c}{17.72} & 
\multicolumn{1}{c}{38.26}  & \multicolumn{1}{c}{14.61} & 
\multicolumn{1}{c}{31.79} & \multicolumn{1}{c}{13.71}\\ \hline

\end{tabular}
\label{tab:result}
\vspace{-10pt}
\end{table*}

Figure \ref{fig:distribution_comparison} demonstrates the accuracy of each round for the 3 models under different datasets and data distributions. In each plot, the horizontal axis represents the training rounds, and the vertical axis represents the average accuracy across 10 edges. The colored solid lines show the rolling average of the accuracy across 10 edges with a window size of 10, while the raw accuracy scores are depicted as dotted, semi-transparent lines. 

Table \ref{tab:result} summarizes the results across all distributions, datasets, and test set types, reporting both \textit{AccN} and \textit{DropM}.

\subsection{Overall Evaluation: Robustness of PHE-FL Performance across Different Edge-Level non-IIDness} 

Overall, PHE-FL demonstrates significant advantages on both imbalanced and balanced test sets under varying degrees of non-IIDness. Notably, \textbf{PHE-FL consistently avoids producing the lowest accuracy across  all distributions.} By contrast, OnlyEdge shows the lowest accuracy in all D3 and D4 cases, whereas EdgeCloud yields the lowest accuracy in D1 and D2. 

\textbf{PHE-FL demonstrates robust performance with respect to accuracy drops, consistently securing the second-smallest drop and avoiding the significant fluctuations observed with EdgeCloud}. EdgeCloud exhibited the largest drop in all scenarios. In contrast, PHE-FL consistently achieved the second-smallest drop across all 15 experimental settings. 

This consistent and reliable performance underscores PHE-FL's robustness and effectiveness, making it a superior choice for environments with diverse data distributions and varying degrees of edge-level non-IIDness.

\subsection{Indepth Evaluation: Automatic Aggregation Adjustment of PHE-FL based on Edge-level Non-IIDness} 

As discussed in Section \ref{sec:3-Motivation}, OnlyEdge is expected to perform well in edge-level non-IID settings, while EdgeCloud is anticipated to excel in IID environments. \textbf{Thus, theoretically, in extreme edge-level non-IID settings (D1), no 3-level FL approach should outperform OnlyEdge, and similarly, in extreme edge-level IID settings (D4), no 3-level FL approach should surpass EdgeCloud.} Since IIDness increases from D1 (extreme non-IID) to D4 (extreme IID), as outlined in Section \ref{sec:data-partition}, our results align with this expectation. Table \ref{tab:result} shows that OnlyEdge consistently ranks highest in D1 and at least second-best in D2, but ranks last in the more IID settings of D3 and D4. Conversely, EdgeCloud ranks highest in D4 and at least second-best in D3, but performs worst in the non-IID settings of D1 and D2.

In contrast, PHE-FL demonstrates robustness across all levels of edge-level IIDness, avoiding poor performance and instability in all 15 cases. This adaptability stems from its \textbf{ability to automatically adjusts the extent to which an edge should adopt knowledge from other models based on performance}, represented as dynamic parameter $\alpha$. When the $\alpha$ parameter is set to 0.1, PHE-FL doesn't have personalization and therefore behaves like EdgeCloud, and when the $\alpha$ parameter is 1, it operates as OnlyEdge, without incorporating knowledge from the globally aggregated model.

Our analysis highlights PHE-FL's ability to navigate diverse edge-level IID and non-IID settings by dynamically adjusting its aggregation strategy based on data distribution. Unlike EdgeCloud and OnlyEdge which perform well in their respective ideal scenarios, \textbf{PHE-FL automatically tunes its personalization mechanism to effectively adapt across most edge-level IID and non-IID conditions, delivering robust and reliable performance in the tested scenarios.}

\subsection{Robustness of PHE-FL with Edge-Level Non-IIDness under Varying Data Complexity}
\label{complexityofdata}
In our experiments, we observed that OnlyEdge consistently achieved over 91.37\% accuracy across all 4 MNIST scenarios, both with imbalanced and balanced test sets. Initially, we suspected overfitting and/or majority label prediction. However, the fact that OnlyEdge excelled in balanced test sets, where test samples included labels that were underrepresented in the training set but overrepresented in the test set, indicates that OnlyEdge was not overfitting to the training dataset. Moreover, the high performance of OnlyEdge under the extreme edge-level IID scenario (D4) suggests that it wasn't simply predicting the majority label. Instead, the results point to the homogeneity of MNIST data within the same labels, where similar features across samples allowed OnlyEdge to achieve high accuracy using only a small portion of the training data. Preliminary experiments confirmed this: a CNN model could achieve 100\% accuracy on the test set for label 1 using just 3\% of its data within five epochs.

However, when tested on more complex datasets like F-MNIST and CIFAR10, OnlyEdge struggled, particularly in edge-level IID settings (D4), due to insufficient data to learn the complex features of each label and the lack of knowledge sharing from other edges. Similarly, in edge-level non-IID settings (D1), EdgeCloud faced challenges on F-MNIST and CIFAR10, showing a significant accuracy gap compared to OnlyEdge and PHE-FL. The forced knowledge sharing among edges with different labels hindered EdgeCloud's ability to learn the complex features specific to each label, instead introduced noise. In contrast, PHE-FL exhibited robust performance across diverse datasets with varying levels of complexity, ranking at least second in all scenarios. 

Thus, beyond the explicit edge-level non-IIDness derived from label skewness, implicit edge-level non-IIDness--represented by the complexity of each label or the difficulty in generalizing data samples within the same label (i.e., feature skewness)--also impacts the performance variance of edge-accommodated FL models. Importantly, \textbf{PHE-FL demonstrated robustness under both explicit (label skewness) and implicit (feature skewness) non-IID conditions, adapting effectively to different levels of data complexity.}

\subsection{Robustness of PHE-FL with Edge-Level Non-IIDness Under Varying Hyperparameters}

\begin{figure}[ht!]
  \centering
  \begin{subfigure}{.45\linewidth}
    \centering
    \includegraphics[width=\linewidth]{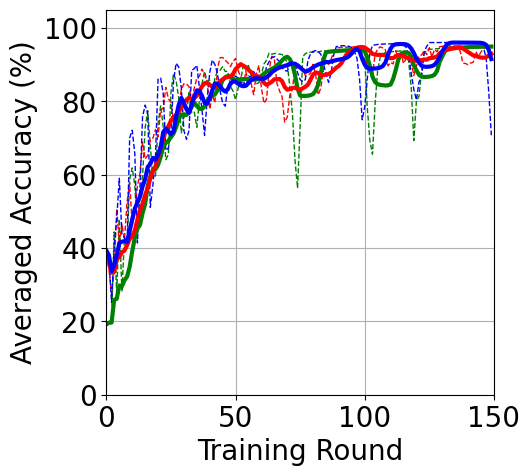}
    \caption{Results of 3 models with 5 devices per edge.}
    \label{fig:scalability_50}
  \end{subfigure}%
  \hfill
  \begin{subfigure}{.45\linewidth}
    \centering
    \includegraphics[width=\linewidth]{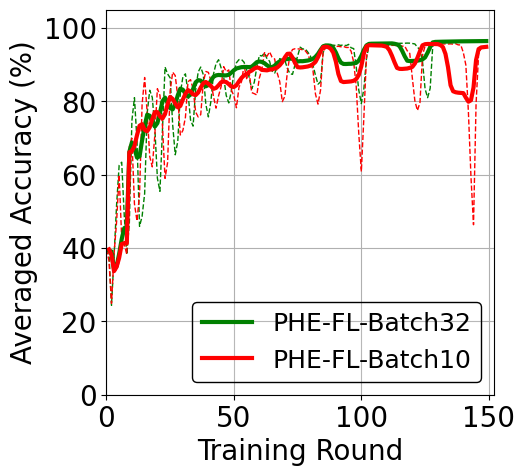}
    \caption{Results of PHE-FL with batch sizes of 32 and 10. }
    \label{fig:batchsize}
  \end{subfigure}
  
  \vspace{10pt} 
  \begin{subfigure}{.45\linewidth} 
    \centering
    \includegraphics[width=\linewidth]{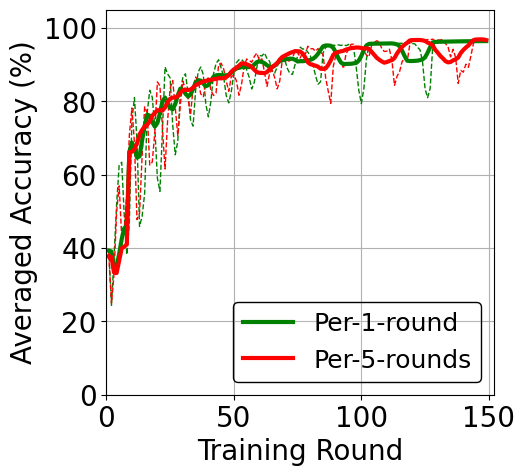}
    \caption{Results of PHE-FL with edge aggregation frequencies of 1 and 5.}
    \label{aggregation_freq}
  \end{subfigure}
  
  \caption{Results of additional experiments testing PHE-FL's robustness with different hyperparameters, all conducted under the MNIST D2 imbalanced test scenario.}
  \vspace{-10pt} 
\end{figure}

We conducted additional experiments under
the MNIST D2 imbalanced test scenario to further validate and generalize PHE-FL's robustness across varying levels of edge-level non-IIDness under different hyperparameters: network size (i.e., different scale), batch size, and aggregation frequency.

First, to test scalability, we used 10 edge nodes, each connected to 5 devices, totaling 50 devices. In this setup, each device had double the data load compared to the experiment with 100 devices, while maintaining the same edge-level non-IID distribution. All other hyperparameters remained the same. Comparing Figure \ref{fig:distribution_comparison} MNIST D2 imbalanced and Figure \ref{fig:scalability_50}, we observed that reducing the number of devices did not significantly impact performance of PHE-FL, suggesting that PHE-FL is resilient to changes in network size.

Next, we evaluated the effect of batch size by reducing it from 32 to 10. As shown in Figure \ref{fig:batchsize}, this change had no notable effect on accuracy or stability, demonstrating PHE-FL’s robustness to batch size variation.

Lastly, we investigated the impact of aggregation frequency by comparing performance when edge-level aggregation occurred every 5 rounds, rather than after every round. Figure \ref{aggregation_freq} shows no significant performance drop, indicating that less frequent aggregation does not adversely affect model performance or stability.

In summary, PHE-FL maintained consistent performance across different network size, batch size, and aggregation frequency settings. These results confirm the robustness of PHE-FL to varying conditions, even when subjected to different hyperparameter configurations, in addition to its inherent ability to perform well under various edge-level non-IIDness scenarios.

%% file: related_works.tex
\section{Related Works}
\vspace{-3pt}
\textbf{Edge-Accommodated 3-level FL:} Enhancing security against inversion attacks \cite{InvertingGradients,ExtraInfoLeakage,ModelInversionAttack} and reducing communication costs \cite{greenfl,CommunicationEfficiency} have been important research tasks in the FL research scene\cite{CESA}. While there are often trade-offs between those two -- enhancing privacy increases communication costs as shown in secure aggregation \cite{SecureAggregation} -- recent studies are on the path to tackle both issues. FedMask \cite{FedMask} introduces ideas to reduce communication costs without breaching but rather enhancing privacy. However, as any form of gradient or update sharing with the cloud can hint at privacy-sensitive data on devices \cite{ExtraInfoLeakage,ModelInversionAttack}, 3-level FL \cite{3-levelFL,EdgeCloud-2}, which does not allow the server to know consecutive updates of individual clients, also seems to be a promising research direction. 

In 3-level FL, IoT devices, edge devices, and the server form a hierarchical structure, enhancing security by concealing device models from the cloud. This framework avoids direct server-client communication, mitigating security risks inherent in 2-level FL \cite{Edge-FL-survey}. Additionally, it reduces communication costs since accessing edge devices is less resource-intensive than the cloud \cite{survey-2}. Hierarchical Federated Averaging (HierFAVG) \cite{EdgeCloud-2} and High-Efficiency and Low-Cost Hierarchical FL (HELCHFL) \cite{3-levelFL} frameworks have demonstrated significant reductions in energy consumption and communication costs compared to traditional 2-level FL. 

\textbf{Non-IIDness handling:} In FL, non-IID  data refers to data distributed with statistical heterogeneity across clients. Skewed feature distribution, skewed label distribution, concept shifting and number of labels are four different factors that result in the varying non-IIDness \cite{non-IID-original}. FL is vulnerable to performance and convergence issues in the presence of non-IID data \cite{TailorFL}. The degradation in performance has been demonstrated in various research papers \cite{GoogleNonIID,FeDDkw}. Given the inherently distributed nature of FL, where data heterogeneity among participating devices is unavoidable, strategic management of non-IID data is essential for sustaining efficiency in these settings. Personalization methods, which tailor models to individual devices, is one of the path to address these challenges.

Personalization strategies for 2-level FL include model mixture, clustering, and pruning. Model Mixture combines a global model with individual local models, as shown in the SuperFed model \cite{measuring-non-iid} and Hierarchical Personalized Federated Learning (HPFL) \cite{HPFL}. Clustering groups of local models based on model similarities, as shown in FedCHAR \cite{HCPFL} and another research \cite{Clustering-2} allows the models for each cluster to converge faster. Pruning adjusts model architectures based on device resources and data heterogeneity, as shown in methods like Tailor FL \cite{TailorFL} FedMask \cite{FedMask} which leveraged sparse binary masks. Despite the useful insights, the FL research on personalization are all focused on 2-level FL. Our research presents a unique approach to addressing hierarchical non-IIDness by shifting the personalization focus from clients to edge nodes, each with distinct optimization goals, within a two-tiered non-IID structure.

%% file: conclusion.tex
\section{Conclusions and Future Work}
In this research, we introduced the concept of hierarchical non-IIDness and highlight that edges often have distinct optimization goals in real-world edge applications. To address edge-level non-IIDness on top of device-level non-IIDness, we developed a novel framework, Personalized Hierarchical Edge-enabled Federated Learning (PHE-FL). PHE-FL enhances model performance at each edge by creating cloud-aggregated models for each edge and using a dynamic parameter $\alpha$ to balance contributions from the edge-aggregated and cloud-aggregated models. Notably, PHE-FL adds no computational burden to client devices and preserves data privacy compared to traditional 2-level FL. We evaluated PHE-FL across 4 distribution scenarios with varying edge-level non-IIDness, showing that it achieves high accuracy and stable robustness compared to existing FL approaches. Our research lays a strong foundation for practical implementation of 3-level FL. Future work will explore deploying PHE-FL in real-world applications and broadening its scope beyond image classification, as well as developing advanced personalization strategies to further enhance its performance, robustness and applicability.